\pdfoutput=1
\documentclass[11pt]{article}

\usepackage{emnlp2021}

\usepackage{times}
\usepackage{latexsym}
\usepackage{amsmath}
\usepackage{amssymb}
\usepackage{graphicx}
\usepackage{multirow}
\usepackage{caption}
\usepackage{subcaption}
\usepackage{diagbox}
\usepackage{amsfonts}
\usepackage[inline]{enumitem}
\usepackage{booktabs,dcolumn}
\usepackage{comment}
\usepackage{xspace}
\usepackage{xcolor}
\usepackage{stfloats}
\usepackage{caption,subcaption}

\usepackage{pifont}
\newcommand{\cmark}{\ding{51}}
\newcommand{\xmark}{\ding{55}}
\usepackage{url}
\newcommand{\mask}{\texttt{[MASK]}}

\newcommand{\agent}{\texttt{[AGENT]}}
\newcommand{\visual}[2]{\begin{subfigure}{0.1\textwidth} \centering \frame{\includegraphics[height=1cm]{figs/images_dir/#1-images/#2.jpg}} \end{subfigure}\hfil}

\title{Worst of Both Worlds:\\ Biases Compound in Pre-trained Vision-and-Language Models}

\author{Tejas Srinivasan \\
  University of Southern California \\
  \texttt{tejas.srinivasan@usc.edu} \\\And
  Yonatan Bisk \\
  Carnegie Mellon University \\
  \texttt{ybisk@cs.cmu.edu} \\}

\date{}

\begin{document}
\maketitle
\begin{abstract}
Numerous works have analyzed biases in vision and pre-trained language models individually - however, less attention has been paid to how these biases interact in multimodal settings. This work extends text-based bias analysis methods to investigate multimodal language models, and analyzes intra- and inter-modality associations and biases learned by these models. Specifically, we demonstrate that VL-BERT~\cite{su2019vl} exhibits gender biases, often preferring to reinforce a stereotype over faithfully describing the visual scene. We demonstrate these findings on a controlled case-study and extend them for a larger set of stereotypically gendered entities.
\end{abstract}

\section{Introduction}
\begin{comment}

\begin{itemize}
    \item Several methods for analyzing bias in text models exist; we extend these methods for VL
    \item In our case study we find that..
\end{itemize}
\end{comment}

Pre-trained contextualized word representations~\cite{peters2018deep, devlin2019bert, radford2018improving, lan2019albert, raffel2020exploring} have been known to amplify unwanted (e.g. stereotypical) correlations from their training data~\cite{zhao2019gender, kurita2019measuring, webster2020measuring, vig2020causal}. By learning these correlations from the data, models may perpetuate harmful racial and gender stereotypes. %Furthermore, models have been known to amplify the biases from their training data, where model output distributions are even more skewed than distributions in the data~\cite{zhao2017men}.

\begin{figure}
    \centering
    \includegraphics[width=\columnwidth]{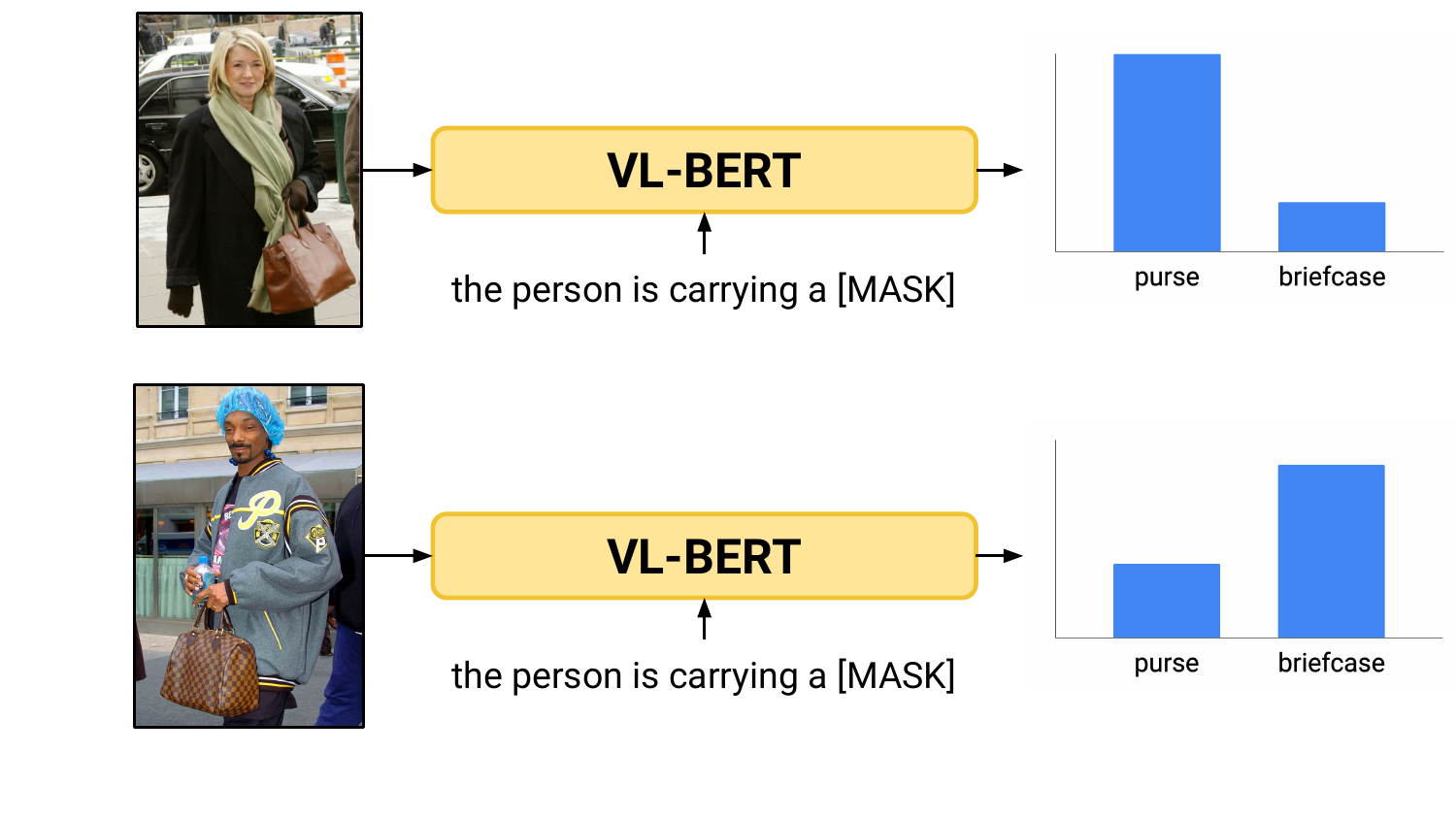}
    \caption{Visual-linguistic models (like VL-BERT) encode gender biases, which (as is the case above) may lead the model to ignore the visual signal in favor of gendered stereotypes.}
    \label{fig:friends-eg}
\end{figure}

The success and generality of pre-trained Transformers has led to several multimodal representation models~\cite{su2019vl, tan2019lxmert, chen2019uniter} which utilize visual-linguistic pre-training. These models also condition on the visual modality, and have shown strong performance on downstream visual-linguistic tasks. This additional input modality allows the model to learn both intra- and inter-modality associations from the training data - and in turn, gives rise to unexplored new sources of knowledge and bias. For instance, we find (see Figure~\ref{fig:friends-eg}) the word \textit{purse}'s female association can override the visual evidence. % might be more strongly associated with images of women rather than men.
While there are entire bodies of work surrounding bias in vision~\cite{buolamwini2018gender} and language~\cite{blodgett-etal-2020-language}, there are relatively few works at the intersection of the two. As we build models that include multiple input modalities, each containing their own biases and artefacts, we must be cognizant about how each of them are influencing model decisions.

In this work, we extend existing work for measuring gender biases in text-only language models to the multimodal setting. Specifically, we study how within- and cross-modality biases are expressed for stereotypically gendered entities in VL-BERT~\cite{su2019vl}, a popular visual-linguistic transformer. Through a controlled case study (\S\ref{sec:case-study}), we find that visual-linguistic pre-training leads to VL-BERT viewing the majority of entities as ``more masculine'' than BERT~\cite{devlin2019bert} does. Additionally, we observe that model predictions rely heavily on the gender of the agent in both the language and visual contexts. These findings are corroborated by an analysis over a larger set of gendered entities (\S\ref{sec:large-scale}).

\section{Bias Statement}
We define gender bias as undesirable variations in how the model associates an entity with different genders, particularly when they reinforce harmful stereotypes.\footnote{In this work, we use ``male" and ``female" to refer to historical definitions of gender presentation. We welcome recommendations on how to generalize our analysis to a more valid characterization of gender and expression.} Relying on stereotypical cues (learned from biased pre-training data) can cause the model to override visual and linguistic evidence when making predictions. This can result in representational harms~\cite{blodgett-etal-2020-language} by perpetuating negative gender stereotypes - \textit{e.g.} men are not likely to hold purses (Figure~\ref{fig:friends-eg}), or women are more likely to wear aprons than suits. In this work, we seek to answer two questions: a) to what extent does visual-linguistic pre-training shift the model's association of entities with different genders? b) do gendered cues in the visual and linguistic inputs \footnote{We note that this work studies biases expressed by models for English language inputs.} influence model predictions?

\section{Methodology}

\begin{comment}
    \item We investigate different sources of bias
    \item general methodology for computing S(E, g) in each source
    \item
\end{comment}

\begin{table*}[t]
\small
\centering
\begin{tabular}{l c l c c c}

     \toprule
     \vspace{1mm} & \multicolumn{2}{c}{To compute $P(E|g)$} & \multicolumn{2}{c}{To compute $P(E|g_N)$} & \\ 
     Source $X$ & \multicolumn{1}{c}{Visual Input} & \multicolumn{1}{c}{Language Input}    & \begin{tabular}[c]{@{}c@{}}Modified\\Component\end{tabular} & \multicolumn{1}{c}{New Value} & \begin{tabular}[c]{@{}c@{}}Association\\Score $S(E,g)$\end{tabular}\\
     \cmidrule{1-1} \cmidrule(lr){2-3} \cmidrule(lr){4-5} \cmidrule{6-6} \vspace{1mm}
     \begin{tabular}[c]{@{}l@{}}Visual-Linguistic\\ Pre-training\end{tabular} & \textcolor{red}{\xmark} & The \texttt{man} is drinking \texttt{beer} & Model & Text-only LM & $\ln \frac{P_{VL}(E | g)}{P_{L}(E | g)}$ \\
     \vspace{1mm}
     \begin{tabular}[c]{@{}l@{}}Language Context\end{tabular} & \begin{subfigure}{0.1\textwidth} \centering \frame{\includegraphics[height=0.8cm]{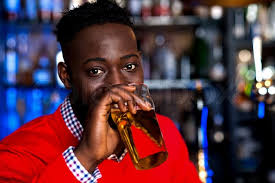}} \end{subfigure} & The \texttt{man} is drinking \texttt{beer} & \begin{tabular}[c]{@{}c@{}}Language\\ Input\end{tabular} & \texttt{man} $\xrightarrow[]{}$ \texttt{person} & $\ln \frac{P_{VL}(E|g, I)}{P_{VL}(E|p, I)}$ \\
     \vspace{1mm}
     \begin{tabular}[c]{@{}l@{}}Visual Context\end{tabular} & \begin{subfigure}{0.1\textwidth} \centering \frame{\includegraphics[height=0.8cm]{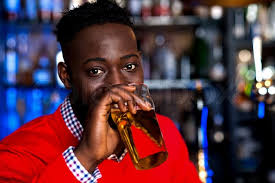}} \end{subfigure} & The \texttt{person} is drinking \texttt{beer} & \begin{tabular}[c]{@{}c@{}}Visual\\ Input\end{tabular} & \textcolor{red}{\xmark} & $\ln \frac{\hat{P}_{VL}(E| I_g)}{P_{VL}(E)}$
     \vspace{-1mm}\\
     
     \bottomrule
     
\end{tabular}
\caption{Our methodology being used to compute association scores $S(E, g)$ between beer ($E$) and man ($g$) in each of the three bias sources. We show the inputs used to compute $P(E|g)$, and the modifications made for the normalizing term, $P(E|g_N)$. 
The entity \texttt{beer} is \mask-ed before being passed into the model.}
\label{tab:methods-table}
\end{table*}

\subsection{Sources of Gender Bias}
\label{subsec:sources}
We identify three sources of learned bias when visual-linguistic models are making masked word predictions - \textbf{visual-linguistic pre-training,} the \textbf{visual context}, and the \textbf{language context}. The former refers to biases learned from image-text pairs during pre-training, whereas the latter two are biases expressed during inference. %For each of these sources, the model is exposed to a new kind of information signal which can induce bias.

\subsection{Measuring Gender Bias}
\label{subsec:measuring}

We measure associations between entities and gender in visual-linguistic models using template-based masked language modeling, inspired by methodology from~\citet{kurita2019measuring}. We provide template captions involving the entity $E$ as language inputs to the model, and extract the probability of the \mask-ed entity. We denote extracted probabilities as:
\begin{align*}
    P_{L/VL}(E | g) & = P(\text{\mask{}}=E | g \text{ in input})
\end{align*}
where $g$ is a gendered agent in one of the input modalities. $L$ and $VL$ are the text-only BERT~\cite{devlin2019bert} and VL-BERT~\cite{su2019vl} models respectively. Our method for computing association scores is simply: 
\begin{align*}
    S(E, g) = \ln \frac{P(E | g)}{P(E|g_N)}
\end{align*}
where the probability terms vary depending on the bias source we want to analyze. We generate the normalizing term by replacing the gendered agent $g$ with a gender-neutral term $g_N$. We summarize how we vary our normalizing term and compute association scores for each bias source in Table~\ref{tab:methods-table}.

\begin{enumerate}
    \item \textbf{Visual-Linguistic Pre-Training ($S_{PT}$):} We compute the association \textit{shift} due to VL pre-training, by comparing the extracted probability $P_{VL}$ from VL-BERT with the text-only BERT - thus $P_L$ is the normalizing term.
    \item \textbf{Language Context ($S_L$):} For an image $I$, we replace the gendered agent $g$ with the gender-neutral term \texttt{person} ($p$) in the caption, and compute the average association score over a set of images $I_E$ which contain the entity $E$.
    \begin{align*}
        S_{L}(E, g) & = \mathbb{E}_{I \sim I_E} \big[ S_{L}(E, g | I) \big]
    \end{align*}
    \item \textbf{Visual Context ($S_V$):} We collect a set of images $I_g$ which contain the entity $E$ and gendered agent $g$, and compute the average extracted probability by providing language input with gender-neutral agent:
    \begin{align*}
    \hat{P}_{VL}(E| I_g) & = \mathbb{E}_{I \sim I_g}[P_{VL}(E | I)]
    \end{align*}
    We normalize by comparing to the output when no image is provided ($P_{VL}(E)$).
\end{enumerate}

For each bias source, we can compute the bias score for that entity by taking the difference of its female and male association scores:
\begin{align*}
    B(E) = S(E, f) - S(E, m)
\end{align*}
The sign of $B(E)$ indicates the direction of gender bias - positive for ``female," negative for ``male."

\section{Case Study}
\label{sec:case-study}
In this section, we present a case study of our methodology by examining how gender bias is expressed in each bias source for several entities. The case study serves as an initial demonstration of our methodology over a small set of gendered entities, whose findings we expand upon in Section~\ref{sec:large-scale}. %We compute the association scores ($S_{PT}, S_V, S_L$) and the gender bias scores for each of these entities.

\subsection{Entities}
\label{subsec:entities}

\begin{table}[t]
\centering
\begin{small}
\begin{tabular}{lll}
\toprule
Template Caption & \multicolumn{2}{c}{Entities} \\ 
\midrule
The \agent{} is carrying a $E$ .  & \textit{purse}&\textit{briefcase}\\ 
The \agent{} is wearing a $E$ . &  \textit{apron}&\textit{suit}\\ 
The \agent{} is drinking $E$ .  & \textit{wine}&\textit{beer}\\ 
\bottomrule
\end{tabular}
\end{small}
\caption{Template captions for each entity pair. The \agent{} is either \textit{man, woman,} or \textit{person} .}
\label{tab:list-templates}
\end{table}

We perform an in-depth analysis of three pairs of entities, each representing a different type of entity: clothes (\textit{apron}, \textit{suit}), bags (\textit{briefcase}, \textit{purse}), and drinks (\textit{wine}, \textit{beer}). The entities are selected to show how unequal gender associations perpetuate undesirable gender stereotypes - \textit{e.g.} aprons are for women, while suits are for men (Appendix~\ref{sec:rationale}). 

For each entity, we collect a balanced set $I_E = I_f \cup I_m$ of 12 images - 6 images each with men ($I_m$) and women ($I_f$) (images in Appendix~\ref{sec:appendix}).\footnote{Note, throughout our discussion we use the words \textit{man} and \textit{woman} as input to the model to denote \textit{male} and \textit{female} to the model. However, when images are included, we only use images of self-identified \textit{(fe)male} presenting individuals.} We also create a different template caption for each entity pair (Table~\ref{tab:list-templates}), which are used to compute association scores $S(E, m/f)$ when the gendered agent $g$ in the caption is \textit{man} or \textit{woman}.  
% For instance, we compute association scores with \textit{male} as:
% \begin{align*}
%     S_{PT}(E, m) & =  \ln \frac{P_{VL}(E | \texttt{[AGENT]}\text{=~man})}{P_{L\phantom{V}}(E | \texttt{[AGENT]}\text{=~man})\hfill} \\
%     S_{V}(E, m) & = \ln \frac{\hat{P}_{VL}(E| \texttt{[AGENT]}\text{=~person}, I_m)}{P_{VL}(E | \texttt{[AGENT]}\text{=~person})\hfill} \\
%     S_{L}(E, m| I) & = \ln \frac{P_{VL}(E|\texttt{[AGENT]}\text{=~man},\phantom{on} I)\hfill}{P_{VL}(E|\texttt{[AGENT]}\text{=~person}, I)\hfill}
% \end{align*}

% \subsection{Models}
% For our visual-linguistic pre-trained language model, we use the base version of VL-BERT~\cite{su2019vl}, which is based on the Transformer architecture, and takes both text and visual inputs in a single stream. When we are computing masked probabilities with text-only inputs i.e. $P_{VL}(E|u)$, we simply omit the visual inputs. When we also have an additional visual input, we extract 100 object proposals using Detectron~\cite{Detectron2018} - the image and object proposals serve as visual input to VL-BERT.
% % 
% For comparisons to a text-only language model, we use the base version of BERT~\cite{devlin2019bert} with the pre-trained weights from Huggingface% Transfomers library
% ~\cite{wolf2019huggingface}. VL-BERT is initialized using BERT's parameters, allowing us to measure association shifts caused by visual-linguistic pretraining.

In the following sections, we analyze how VL-BERT exhibits gender bias for these entities, for each of the bias sources identified in Section~\ref{subsec:sources}.

\begin{figure}[t]
    \centering
    \includegraphics[width=\columnwidth]{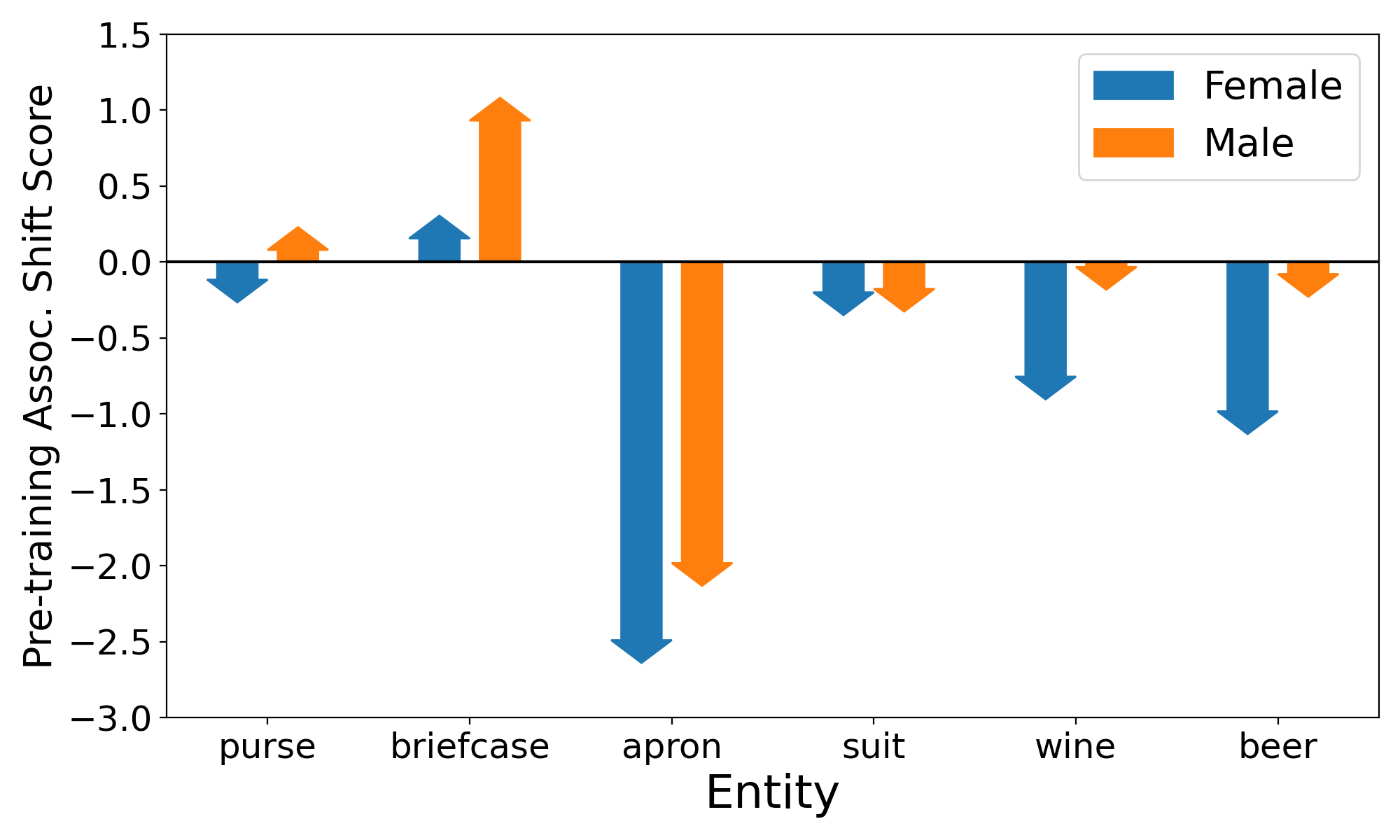}
    \caption{Pre-training association shift scores $S_{PT}(E, m/f)$. Positive shift scores indicate that VL-BERT has higher associations between the entity and the agent's gender than BERT, and vice versa}
    \label{fig:ptbias}
\end{figure}

\subsection{Visual-Linguistic Pre-Training Bias}
\label{subsec:casestudy-pt}

Figure~\ref{fig:ptbias} plots each entity's pre-training association shift score, $S_{PT}(E, m/f)$, where positive scores indicate that visual-linguistic pre-training amplified the gender association, and vice versa. %The difference between \textit{female} and \textit{male} association shift scores represents the entity's gender bias caused by visual-linguistic pre-training, $B_{PT}(E)$. 

Visual-linguistic pre-training affects all objects differently. Some objects have increased association scores for both genders (\textit{briefcase}), while others have decreased associations (\textit{suit} and \textit{apron}). However, even when the associations shift in the same direction for both genders, they rarely move together - for \textit{briefcase}, the association increase is much larger for male, whereas for \textit{apron, wine} and \textit{beer}, the association decrease is more pronounced for female. %The exception is \textit{suit}, for which both association shift scores are approximately the same. 
For \textit{purse}, the association shifts positively for male but negatively for female. For the entities in the case study, we conclude that pretraining shifts entities' association towards men.

\subsection{Language Context Bias}
\label{subsec:casestudy-l}

\begin{figure}[b]
    \centering
    \includegraphics[width=\columnwidth]{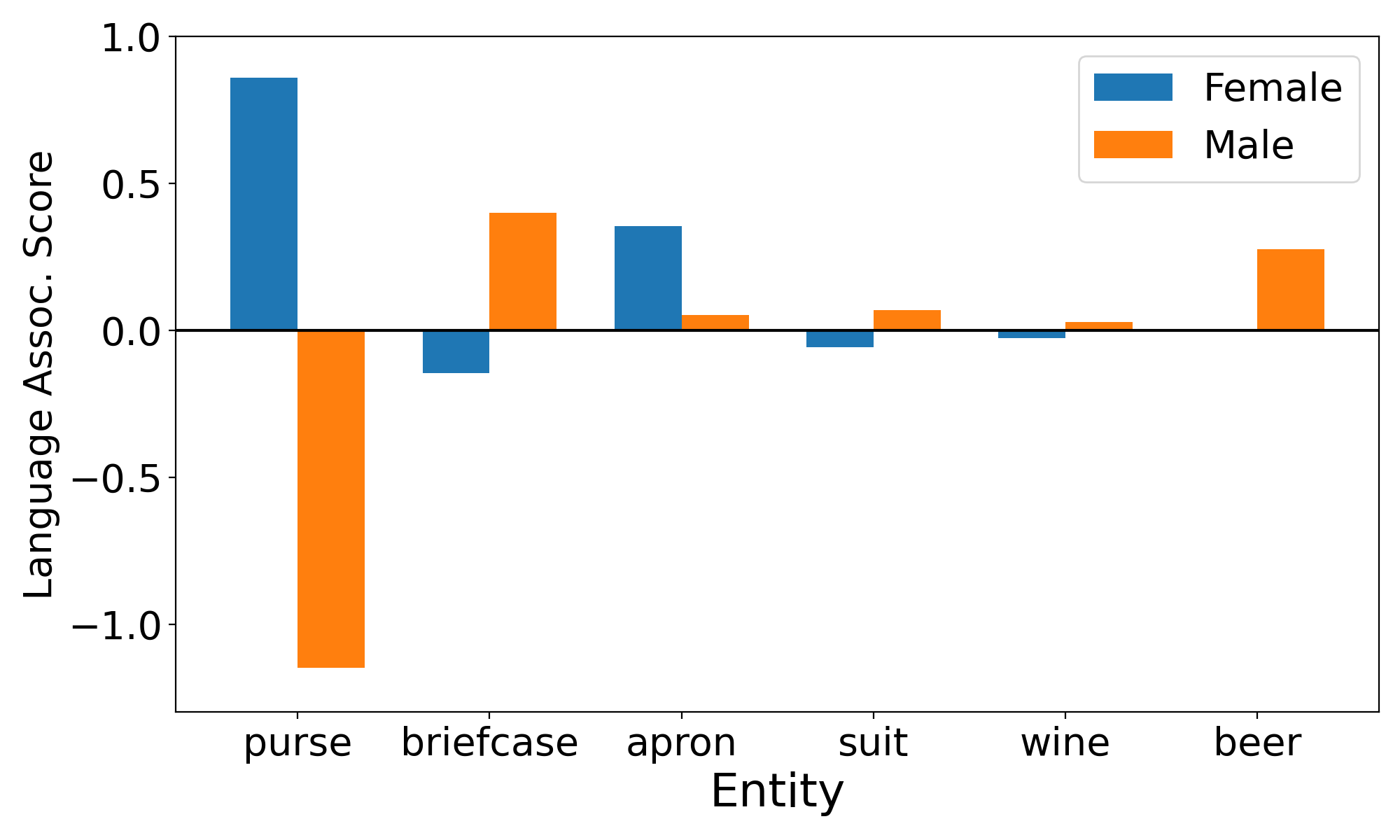}
    \caption{Language association scores $S_L(E, m/f)$. Positive association scores indicate that the agent's gender increases the model's confidence in the entity.}
    \label{fig:lbias}
\end{figure}

Figure~\ref{fig:lbias} plots language association scores, which look at the masked probability of $E$ when the agent in the caption is \textit{man/woman}, compared to the gender-neutral \textit{person}. 

For the entity \textit{purse}, we see that when the agent in the language context is female the model is much more likely to predict that the masked word is \textit{purse}, but when the agent is male the probability becomes much lower. We similarly observe that some of the entities show considerably higher confidence when the agent is either male or female (\textit{briefcase, apron, beer}), indicating that the model has a language gender bias for these entities. For \textit{suit} and \textit{wine}, association scores with both genders are similar.

\subsection{Visual Context Bias}

\begin{figure}[t]
    \centering
    \includegraphics[width=\columnwidth]{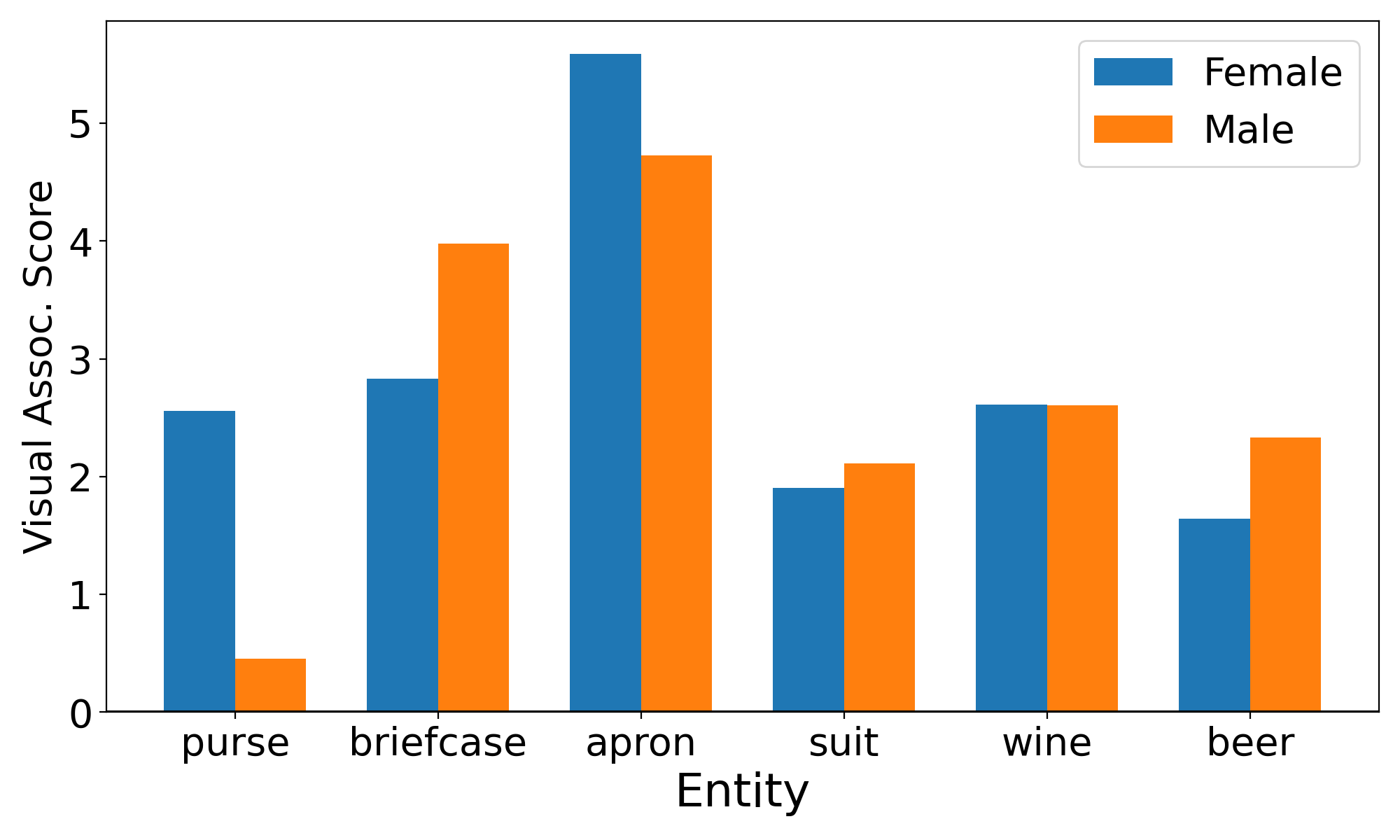}
    \caption{Visual association scores $S_V(E, m/f)$. Positive association scores indicate that the model becomes more confident in the presence of a visual context.}
    \label{fig:vbias}
\end{figure}

For each of our entities, we also plot the visual association score $S_V(E, u)$ with \textit{male} and \textit{female} in Figure~\ref{fig:vbias}. We again observe that the degree of association varies depending on whether the image contains a man or woman. For \textit{purse} and \textit{apron}, the model becomes considerably more confident in its belief of the correct entity when the agent is female rather than male. Similarly, if the agent is male, the model becomes more confident about the entity in the case of \textit{briefcase} and \textit{beer}. For \textit{suit} and \textit{wine}, the differences are not as pronounced. In Table~\ref{tab:examples}, we can see some examples of the model's probability outputs not aligning with the object in the image. In both cases, the model's gender bias overrides the visual evidence (the entity).

% The main takeaways from our case study are:
% \begin{itemize}
%     \itemsep0em
%     \item Visual-linguistic pre-training enhances entity associations more for men, and dampens associations more for women
%     \item Model decisions are unequally influenced by the gender in the language and visual context
% \end{itemize}

\section{Comparing Model Bias with Human Annotations of Stereotypes} %Beyond the Case Study: Bias Analysis on Larger Set of Entities}
\label{sec:large-scale}
To test if the trends in the case study match human intuitions, we 
%In order to show that the trends captured in the case study are more general and not just an artefact of the entities we handpicked, we also conduct a bias analysis over a larger set of gendered entities. We 
curate a list of 40 entities, which are considered to be stereotypically masculine or feminine in society.\footnote{We surveyed 10 people and retained 40/50 entities where majority of surveyors agreed with a stereotyped label.} We analyze how the gendered-ness of these entities is mirrored in their VL-BERT language bias scores. To evaluate the effect of multimodal training on the underlying language model, 
%observe how biases from visual-linguistic pre-training are reflected in the language model, 
we remove the visual input when extracting language model probabilities and %We further 
compare how the language bias varies between text-only VL-BERT and the text-only BERT model.

\begin{table}[t]
\centering
\begin{tabular}{c c c}

     \toprule
     \multirow{1}{*}{Visual Context, $I$} & \frame{\includegraphics[width=1.5cm]{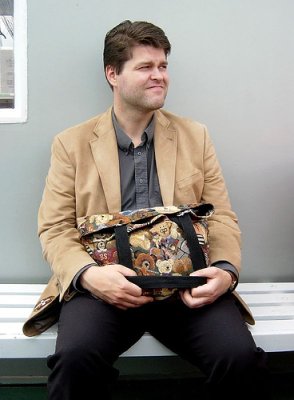}} & \frame{\includegraphics[width=1.5cm]{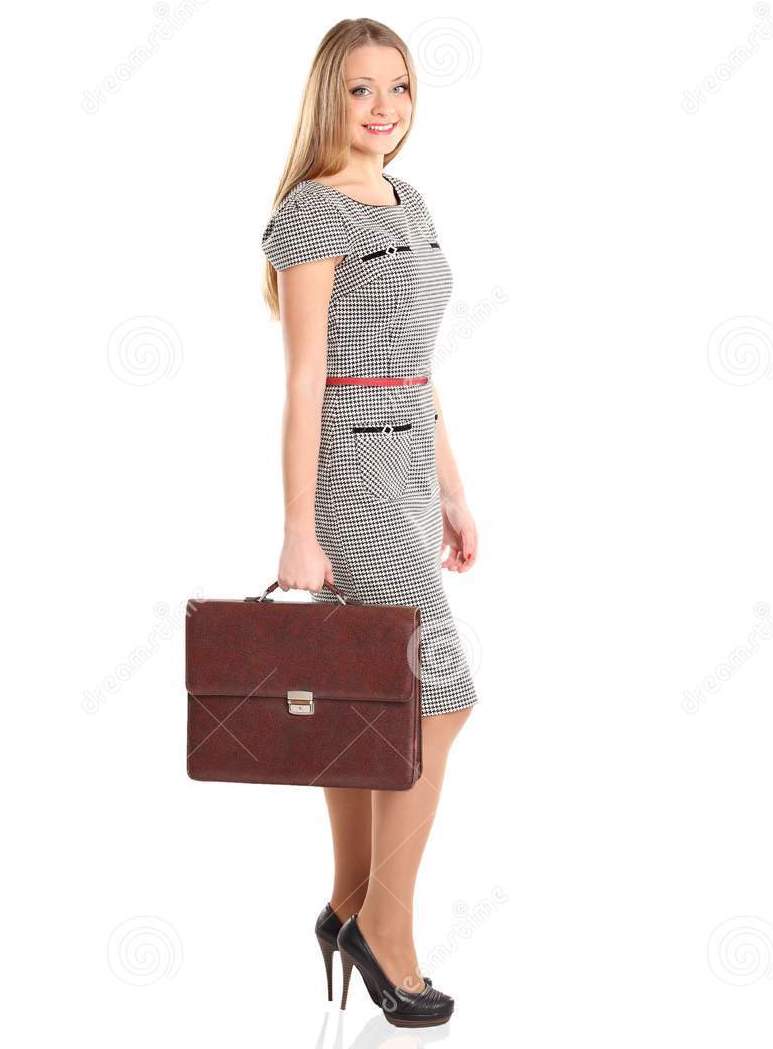}} \\ 
     \midrule \\& & \vspace{-10 mm} \\
     $P_{VL}(\text{purse} | I)$ &     $0.0018$ \textcolor{green}{\cmark} &      $0.084$ \textcolor{red}{\xmark} \\
     $P_{VL}(\text{briefcase} | I)$ &     $0.4944$ \textcolor{red}{\xmark} &     $0.067$ \textcolor{green}{\cmark} \\
     \bottomrule
     
\end{tabular}
\caption{Examples of images where the probability outputs do not align with the visual information.}
\label{tab:examples}
\end{table}

\begin{figure*}[htp]
\centering
\subfloat[$B_L^{VLBert}$ for 40 entities which are stereotypically considered \textcolor{orange}{masculine} or \textcolor{blue}{feminine}. For the majority of entities, the direction of the gender bias score aligns with the stereotypical gender label, indicating that VL-BERT reflects these gender stereotypes.]{%
  \label{subfig:bias-vlbert}
  \includegraphics[clip,width=\linewidth]{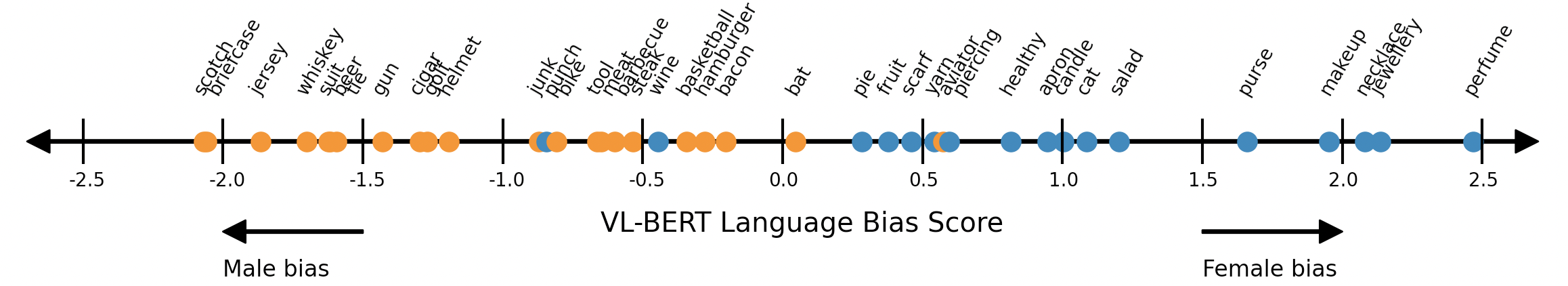}%
}

\subfloat[$B_L^{VLBert}(E) - B_L^{Bert}(E)$ for the 40 gendered entities. The distribution of entities is skewed towards increased masculine/decreased feminine association for VL-BERT, indicating VL pre-training shifts the association distribution for most entities towards men. Note that VL-BERT still associates \textit{cat} with women and \textit{cigar} with men (see~\ref{subfig:bias-vlbert}), but less strongly than BERT.]{%
  \label{subfig:bias-diff}
  \includegraphics[clip,width=\linewidth]{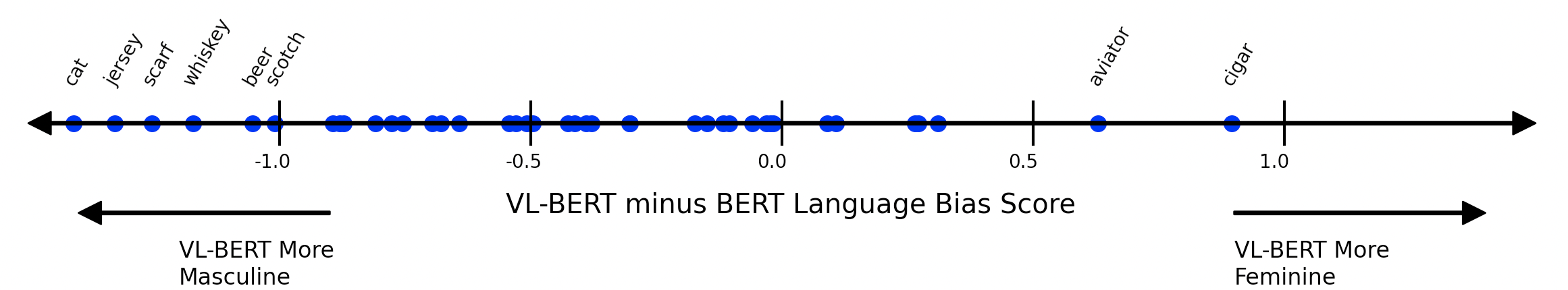}%
}
\caption{}
\label{fig:large-scale}
\end{figure*}

For the language input, we create template captions similar to those described in Table~\ref{tab:list-templates}. For every entity $E$, we compute the language bias score $B_L(E)$ by extracting probabilities from the visual-linguistic model, $P_{VL}(E|f/m/p)$.
\begin{align*}
    S_L(E, m/f) & = \ln \frac{P_{VL}(E|m/f)}{P_{VL}(E|p)}\\
    B_{L}^{VLBert}(E) & = S_L(E,f) - S_L(E,m) \\
    & = \ln \frac{P_{VL}(E|f)}{P_{VL}(E|m)}
\end{align*}
Positive values of $B_{VL}(E)$ correspond to a female bias for the entity, while negative values correspond to a male bias. We plot the bias scores in Table~\ref{subfig:bias-vlbert}. We see that the language bias scores in VL-BERT largely reflect the stereotypical genders of these entities - indicating that the results of Section~\ref{subsec:casestudy-l} generalize to a larger group of entities.

We can also investigate the effect of visual-linguistic pretraining by comparing these entities' VL-BERT gender bias scores with their gender bias scores under BERT. We compute the language bias score for BERT, $B_L^{Bert}(E)$, by using the text-only language model probability $P_L(E|g)$ instead. We plot the difference between entities' VL-BERT and BERT bias scores in Table~\ref{subfig:bias-diff}. Similar to trends observed in Section~\ref{subsec:casestudy-pt}, we see that the majority of objects have increased masculine association after pre-training ($B_{L}^{VLBert} < B_L^{Bert}$).

\section{Related Work}
\begin{comment}

\begin{itemize}
    \item Pre-trained L+V Language Models
    \begin{itemize}
        \item Pre-trained LMs: ELMo, BERT
        \item Visual linguistic LMs: VL-BERT, ViLBERT, LXMERT
    \end{itemize}
    \item Bias Quantification in LMs/Embeddings
    \begin{itemize}
        \item \cite{kurita2019measuring}
        \item \cite{zhao2019gender}
    \end{itemize}
    \item Bias Amplification?
    \begin{itemize}
        \item \cite{zhao2017men}
    \end{itemize}
\end{itemize}
\end{comment}

\paragraph{Vision-and-Language Pre-Training} %Recent attempts to bridge vision and language, and the advent of large pre-trained language representations, have led to several efforts to build joint vision-and-language representation models. These representation models are built on the Transformer~\cite{vaswani2017attention} backbone. 
Similar to BERT~\cite{devlin2019bert}, vision-and-language transformers  \cite{su2019vl,tan2019lxmert,chen2019uniter} are trained with masked language modeling and region modeling with multiple input modalities. 
%Other models, like ViLBERT~\cite{Lu_2020_CVPR}, are pre-trained on multiple vision-and-language tasks. 
These models yield state-of-the-art results on many multimodal tasks: e.g. VQA~\cite{antol2015vqa}, Visual Dialog~\cite{das2017visual}, and VCR~\cite{zellers2019recognition}.% and Vision-and-Language Navigation~\cite{anderson2018vision}.

\paragraph{Bias Measurement in Language Models}
%Much work has studied the existence and measurement of gender bias in language models. 
~\citet{bolukbasi2016man} and ~\citet{caliskan2017semantics} showed that static word embeddings like Word2Vec and GloVe encode biases about gender roles. 
%Much work has also studied the downstream effect of these 
Biases negatively effect downstream tasks (e.g. coreference \cite{zhao2018gender, rudinger2018gender}) and exist in large pretrained models 
%As the community moved towards contextual word representations like ELMo~\cite{peters2018deep} and BERT~\cite{devlin2019bert}, several methods have been proposed to measure gender bias in these representations
~\cite{zhao2019gender, kurita2019measuring, webster2020measuring}. Our methodology is inspired by ~\citet{kurita2019measuring}, who utilized templates and the Masked Language Modeling head of BERT to show how different probabilities are extracted for different genders. We extend their text-only methodology to vision-and-language models.

\paragraph{Bias in Language + Vision} Several papers have investigated how dataset biases can override visual evidence in model decisions. \citet{zhao2017men} showed that multimodal models can amplify gender biases in training data. In VQA, models make decisions by exploiting language priors rather than utilizing the visual context~\cite{goyal2017making, ramakrishnan2018overcoming}. Visual biases can also affect language, where gendered artefacts in the visual context influence generated captions~\cite{hendricks2018women, bhargava2019exposing}.
%\newpage 

\section{Future Work and Ethical Considerations}
This work extends the bias measuring methodology of ~\citet{kurita2019measuring} to multimodal language models. Our case study shows that these language models are influenced by gender information from both language and visual contexts - often ignoring visual evidence in favor of stereotypes.

%There are several important directions for future research. While our current gender bias analysis is restricted to the binary setting, searching for biases in a more representative pool of the population requires self-identification for images of gender (including but not limited to gender non-conforming, non-binary, and trans individuals). Similarly questions around race and skin-color can be investigated \textit{if} appropriate prompts can be crafted. Future efforts should aim to mitigate the biases identified here. For example, models, like humans, should not guess at gender identity, nor should they rely on tropes in lieu of visual evidence. %For instance, if our model is learning biases due to gender imbalances in the pretraining dataset(s), would these biases go away if we pretrain the multimodal models using a balanced dataset?

Gender is not binary, but this work performs bias analysis for the terms ``male" and ``female" -- which are traditionally proxies for cis-male and cis-female.  In particular, when images are used of male and female presenting individuals we use images that self-identify as male and female.  We avoid guessing at gender presentation and note that the biases studied here in this unrealistically simplistic treatment of gender pose even more serious concerns for gender non-conforming, non-binary, and trans-sexual individuals.  A critical next step is designing more inclusive probes, and training (multi-modal) language models on more inclusive data. We welcome criticism and guidance on how to expand this research.  Our image based data suffers from a second, similar, limitation on the dimension of race. All individuals self-identified as ``white" or ``black", but a larger scale inclusive data-collection should be performed across cultural boundaries and skin-tones with the self-identification and \textit{if} appropriate prompts can be constructed for LLMs.

%\section*{Ethical Considerations}
%\input{ethical}

%\bibliography{acl2018}
\bibliography{acl2021_new}
\bibliographystyle{acl_natbib}

\clearpage
\appendix
\section{Images Collected for Case Study}

In Table~\ref{tab:images_used}, we show the different images collected for our Case Study in Section~\ref{sec:case-study}.
\label{sec:appendix}
\begin{table*}[ht]
\centering
\begin{tabular}{c c c}
     \toprule
     Entity & Gender of Agent & Images Used ($I_{m/f}$) \\ 
     \midrule 
     \multirow{2}{*}{Purse} & Male & 
    \visual{purse}{1} \visual{purse}{2} \visual{purse}{3} \visual{purse}{4} \visual{purse}{5} \visual{purse}{6} \\
    \cmidrule{2-3} 
     & Female & 
    \visual{purse}{7} \visual{purse}{8} \visual{purse}{9} \visual{purse}{10} \visual{purse}{11} \visual{purse}{12}\\
    \midrule 
     \multirow{2}{*}{Briefcase} & Male & 
    \visual{briefcase}{1} \visual{briefcase}{2} \visual{briefcase}{3} \visual{briefcase}{4} \visual{briefcase}{5} \visual{briefcase}{6} \\
    \cmidrule{2-3} 
     & Female & 
    \visual{briefcase}{7} \visual{briefcase}{8} \visual{briefcase}{9} \visual{briefcase}{10} \visual{briefcase}{11} \visual{briefcase}{12}\\
    \midrule 
     \multirow{2}{*}{Apron} & Male & 
    \visual{apron}{1} \visual{apron}{2} \visual{apron}{3} \visual{apron}{4} \visual{apron}{5} \visual{apron}{6} \\
    \cmidrule{2-3} 
     & Female & 
    \visual{apron}{7} \visual{apron}{8} \visual{apron}{9} \visual{apron}{10} \visual{apron}{11} \visual{apron}{12}\\
    \midrule 
     \multirow{2}{*}{Suit} & Male & 
    \visual{suit}{1} \visual{suit}{2} \visual{suit}{3} \visual{suit}{4} \visual{suit}{5} \visual{suit}{6} \\
    \cmidrule{2-3} 
     & Female & 
    \visual{suit}{7} \visual{suit}{8} \visual{suit}{9} \visual{suit}{10} \visual{suit}{11} \visual{suit}{12}\\
    \midrule 
     \multirow{2}{*}{Wine} & Male & 
    \visual{wine}{1} \visual{wine}{2} \visual{wine}{3} \visual{wine}{4} \visual{wine}{5} \visual{wine}{6} \\
    \cmidrule{2-3} 
     & Female & 
    \visual{wine}{7} \visual{wine}{8} \visual{wine}{9} \visual{wine}{10} \visual{wine}{11} \visual{wine}{12}\\
    \midrule 
     \multirow{2}{*}{Beer} & Male & 
    \visual{beer}{1} \visual{beer}{2} \visual{beer}{3} \visual{beer}{4} \visual{beer}{5} \visual{beer}{6} \\
    \cmidrule{2-3} 
     & Female & 
    \visual{beer}{7} \visual{beer}{8} \visual{beer}{9} \visual{beer}{10} \visual{beer}{11} \visual{beer}{12}\\
     \bottomrule
     
\end{tabular}
\caption{Images collected for case study in Section 4}
\label{tab:images_used}
\end{table*}

\section{Rationale Behind Selection of Case Study Entities}
\label{sec:rationale}

For the purpose of the case study, we chose three pairs of entities, each containing entities with opposite gender polarities (verified using the same survey we used in Section~\ref{sec:large-scale}). The entities were chosen to demonstrate how unequal gender associations perpetuate undesirable gender stereotypes.

\paragraph{Apron vs Suit} This pair was chosen to investigate how clothing biases can reinforce stereotypes about traditional gender roles. Aprons are associated with cooking, which has long been considered a traditional job for women as homemakers. Meanwhile, suits are associated with business, and men are typically considered to be the breadwinners for their family. However, in the 21st century, as we make progress in breaking the breadmaker-homemaker dichotomy, these gender roles do not necessarily apply~\cite{cunningham2008changing, zuo2000breadwinner}, and reinforcing them is harmful - particularly to women, since they have struggled (and continue to struggle) for their right to join the workforce and not be confined by their gender roles.

\paragraph{Purse vs Briefcase} Bags present another class of traditional gender norms that are frequently violated in this day and age. Purses are traditionally associated with women, whereas briefcases (similar to suits above) are associated with business, which we noted is customarily a male occupation. If a model tends to associate purses with women, in the presence of contrary visual evidence, it could reinforce heteronormative gender associations. Similarly, associating briefcases with primarily men undermines the efforts of women to enter the workforce.

\paragraph{Wine vs Beer} Alcoholic drinks also contain gendered stereotypes that could be perpetuated by visual-linguistic models. Beer is typically considered to be a masculine drink~\cite{fugitt2018beer, darwin2018omnivorous}, whereas wine is associated with feminine traits~\cite{landrine1988gender}.

% \section{Examples of VCR Rationales}
% \label{sec:vcr-examples}
% In Table~\ref{tab:vcr-examples}, we present several examples from the VCR rationales, used in Section~\ref{subsec:realworld}.
% \input{vcr_examples}

\end{document}